\documentclass[11pt]{article}
\usepackage{graphics}
\usepackage{amsmath}
\usepackage{amssymb}

\newcommand{\m}{{\bf m}}
\newcommand{\h}{{\bf h}}

\newcommand{\V}{{\bf V}}
\newcommand{\bH}{{\bf H}}

\newcommand{\W}{{\bf W}}

\newcommand{\x}{{\bf x}}
\newcommand{\s}{{\bf s}}

\newcommand{\wi}{{\bf w}_i}
\newcommand{\hi}{{\bf h}_i}

\newcommand{\vv}{{\bf v}}

\newcommand{\beq}{\begin{equation}}
\newcommand{\eeq}{\end{equation}}

\begin{document}

\title{\bf Non-negative matrix factorization with\\ sparseness constraints}

\author{Patrik O.\ Hoyer\\[4mm]
HIIT Basic Research Unit\\
Department of Computer Science\\
University of Helsinki, Finland\\
www.cs.helsinki.fi/patrik.hoyer\\
}
\date{\vspace{1mm} Submitted manuscript,  \today}

\maketitle


\begin{abstract}
Non-negative matrix factorization (NMF) is a recently developed technique for finding parts-based, linear representations of non-negative data. Although it has successfully been applied in several applications, it does not always result in parts-based representations. In this paper, we show how explicitly incorporating the notion of `sparseness' improves the found decompositions. Additionally, we provide complete MATLAB code both for standard NMF and for our extension. Our hope is that this will further the application of these methods to solving novel data-analysis problems.
\end{abstract}
\vspace{3mm}

\centerline{
}


\section{Introduction}
\label{sec:intro}

A fundamental problem in many data-analysis tasks is to find a suitable representation of the data. A useful representation typically makes latent structure in the data explicit, and often reduces the dimensionality of the data so that further computational methods can be applied. 

Non-negative matrix factorization (NMF) \cite{Paatero94,LeeDD99} is a recent method for finding such a representation. Given a non-negative data matrix $\V$, NMF finds an approximate factorization $\V \approx \W\bH$ into non-negative factors $\W$ and $\bH$. The non-negativity constraints make the representation purely additive (allowing no subtractions), in contrast to many other linear representations such as principal component analysis (PCA) and independent component analysis (ICA) \cite{Hyva01book}. 

One of the most useful properties of NMF is that it usually produces a \emph{sparse} representation of the data. Such a representation encodes much of the data using few `active' components, which makes the encoding easy to interpret. Sparse coding \cite{Field94} has also, on theoretical grounds, been shown to be a useful middle ground between completely distributed representations, on the one hand, and unary representations (grandmother cells) on the other \cite{Foldiak95,Thorpe95}. However, because the sparseness given by NMF is somewhat of a side-effect rather than a goal, one cannot in any way control the degree to which the representation is sparse. In many applications, more direct control over the properties of the representation is needed.

In this paper, we extend NMF to include the option to control sparseness explicitly. We show that this allows us to discover parts-based representations that are qualitatively better than those given by basic NMF. We also discuss the relationship between our method and other recent extensions of NMF \cite{LiS01,Hoyer02NNSP,Liu03}.

Additionally, this contribution includes a complete MATLAB package for performing NMF and its various extensions. Although the most basic version of NMF requires only two lines of code and certainly does not warrant distributing a separate software package, its several extensions involve more complicated operations; the absense of ready-made code has probably hindered their widespread use so far. We hope that our software package will alleviate the problem.

This paper is structured as follows. In section~\ref{sec:nmf} we describe non-negative matrix factorization, and discuss its success but also its limitations. Section~\ref{sec:nmfsparse} discusses why and how to incorporate sparseness constraints into the NMF formulation. Section~\ref{sec:experiments} provides experimental results that verify our approach. Finally, sections~\ref{sec:discussion} and \ref{sec:conclusions} compare our approach to other recent extensions of NMF and conclude the paper.


\section{Non-negative matrix factorization}
\label{sec:nmf}

Non-negative matrix factorization is a \emph{linear}, \emph{non-negative} approximate data representation. Let's assume that our data consists of $T$ measurements of $N$ non-negative scalar variables. Denoting the ($N$-dimensional) measurement vectors $\vv^t$ ($t=1,\ldots, T$), a linear approximation of the data is given by
\beq
\vv^t \approx \sum_{i=1}^M \wi h_i^t = \W\h^t,
\eeq
where $\W$ is an $N\times M$ matrix containing the \emph{basis vectors} $\wi$ as its columns. Note that each measurement vector is written in terms of the \emph{same} basis vectors. The $M$ basis vectors $\wi$ can be thought of as the `building blocks' of the data, and the ($M$-dimensional) coefficient vector $\h^t$ describes how strongly each building block is present in the measurement vector $\vv^t$.

Arranging the measurement vectors $\vv^t$ into the columns of an $N\times T$ matrix $\V$, we can now write
\beq
\V \approx \W\bH,
\eeq
where each column of $\bH$ contains the coefficient vector $\h^t$ corresponding to the measurement vector $\vv^t$. Written in this form, it becomes apparent that a linear data representation is simpy a factorization of the data matrix. Principal component analysis, independent component analysis, vector quantization, and non-negative matrix factorization can all be seen as matrix factorization, with different choices of objective function and/or constraints.

Whereas PCA and ICA do not in any way restrict the signs of the entries of $\W$ and $\bH$, NMF requires all entries of both matrices to be non-negative. What this means is that the data is described by using additive components only. This constraint has been motivated in a couple of ways: First, in many applications one knows (e.g.\ by the rules of physics) that the quantities involved cannot be negative. In such cases, it can be difficult to interpret the results of PCA and ICA \cite{Paatero94,Parra00}. Second, non-negativity has been argued for based on the intuition that parts are generally combined additively (and not subtracted) to form a whole; hence, these constraints might be useful for learning parts-based representations \cite{LeeDD99}. 

Given a data matrix $\V$, the optimal choice of matrices $\W$ and $\bH$ are defined to be those non-negative matrices that minimize the reconstruction error between $\V$ and $\W\bH$. Various error functions have been proposed \cite{Paatero94,LeeDD01}, perhaps the most widely used is the squared error (euclidean distance) function
\beq
E(\W,\bH) = \|\V - \W\bH\|^2 = \sum_{i,j} (V_{ij} - (\W\bH)_{ij})^2.
\eeq
Although the minimization problem is convex in $\W$ and $\bH$ separately, it is not convex in both simultaneously. Paatero \cite{Paatero94} gave a gradient algorithm for this optimization, whereas Lee and Seung \cite{LeeDD01} devised a multiplicative algorithm that is somewhat simpler to implement and also showed good performance.

\begin{figure}
\centerline{
\resizebox{100mm}{!}{
\includegraphics{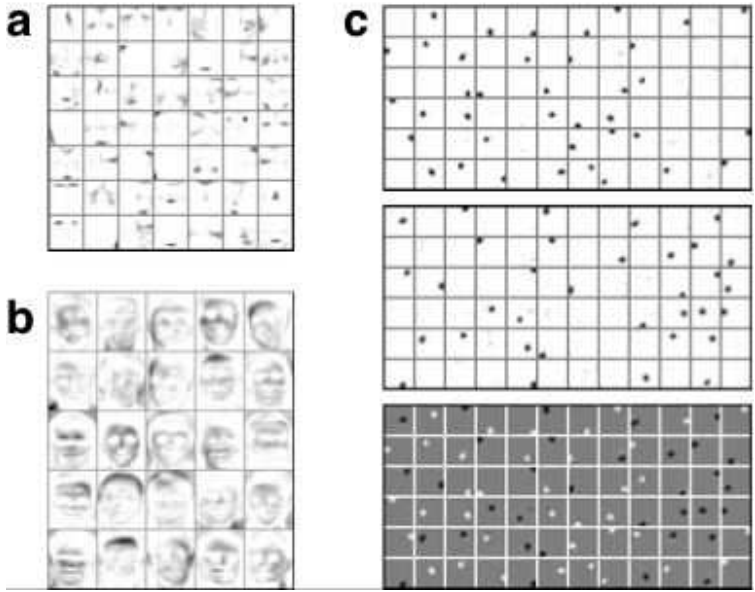}}}
\caption{NMF applied to various image datasets. {\bf (a)} Basis images given by NMF applied to face image data from the CBCL database \cite{CBCLfacedatabase}, following \cite{LeeDD99}. In this case NMF produces a parts-based representation of the data. {\bf (b)} Basis images derived from the ORL face image database \cite{ORLfacedatabase}, following \cite{LiS01}. Here, the NMF representation is global rather than parts-based. {\bf (c)} Basis vectors from NMF applied to ON/OFF-contrast filtered natural image data \cite{Hoyer03CNS}. Top: Weights for the ON-channel. Each patch represents the part of one basis vector $\wi$ corresponding to the ON-channel. (White pixels denote zero weight, darker pixels are positive weights.) Middle: Corresponding weights for the OFF-channel. Bottom: Weights for ON minus weights for OFF. (Here, gray pixels denote zero.) Note that NMF represents this natural image data using circularly symmetric features. 
\label{fig:nmf}}
\end{figure}

Although some theoretical work on the properties of the NMF representation exists \cite{Donoho04}, much of the appeal of NMF comes from its empirical success in learning meaningful features from a diverse collection of real-life datasets. Lee and Seung \cite{LeeDD99} showed that, when the dataset consisted of a collection of face images \cite{CBCLfacedatabase}, the representation consisted of basis vectors encoding for the mouth, nose, eyes, etc; the intuitive features of face images. In Figure~\ref{fig:nmf}a we have reproduced that basic result using the same dataset. Additionally, they showed that meaningful topics can be learned when text documents are used as data. Subsequently, NMF has been successfully applied to a variety of datasets \cite{Buchsbaum02,Brunet04,Jung04,KimPM03}.

Despite this success, there also exist datasets for which NMF does not give an intuitive decomposition into parts that would correspond to our idea of the `building blocks' of the data. In \cite{LiS01}, the authors showed that when NMF was applied to a different facial image database \cite{ORLfacedatabase}, the representation was global rather than local, qualitatively different from that reported in \cite{LeeDD99}. Again, we have rerun that experiment and confirm those results, see Figure~\ref{fig:nmf}b. The difference was mainly attributed to how well the images were hand-aligned \cite{LiS01}. 

Another case where the decomposition found by NMF does not match the underlying elements of the data is shown in figure~\ref{fig:nmf}c. In this experiment \cite{Hoyer03CNS}, natural image patches were high-pass filtered and subsequently split into positive (`ON') and negative (`OFF') contrast channels, in a process similar to how visual information is processed by the retina. When NMF is applied to such a dataset, the resulting decomposition does not consist of the oriented filters which form the cornerstone of most of modern image processing. Rather, NMF represents these images using simple, dull, circular `blobs'.

We will show that, in both of the above cases, explicitly controlling the sparseness of the representation leads to representations that are parts-based and match the intuitive features of the data.


\section{Adding sparseness constraints to NMF}
\label{sec:nmfsparse}

\subsection{Sparseness}
\label{sec:sparseness}

\begin{figure}
\centerline{
\resizebox{130mm}{!}{
\includegraphics{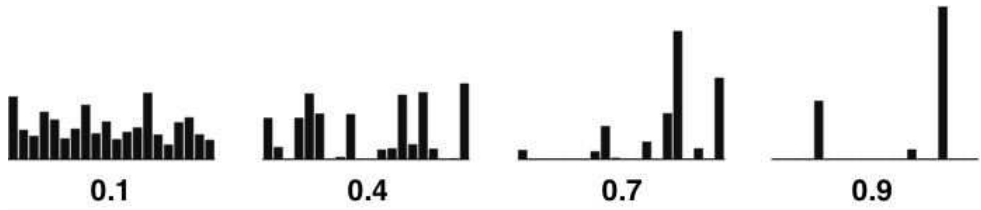}}
}
\caption{Illustration of various degrees of sparseness. Four vectors are shown, exhibiting sparseness levels of 0.1, 0.4, 0.7, and 0.9. Each bar denotes the value of one element of the vector. At low levels of sparseness (leftmost), all elements are roughly equally active. At high levels (rightmost), most coefficients are zero whereas only a few take significant values. 
\label{fig:sparseness}}
\end{figure}

The concept of `sparse coding' refers to a representational scheme where only a few units (out of a large population) are effectively used to represent typical data vectors \cite{Field94}. In effect, this implies most units taking values close to zero while only few take significantly non-zero values. Figure~\ref{fig:sparseness} illustrates the concept and our sparseness measure (defined below).

Numerous sparseness measures have been proposed and used in the literature to date. Such measures are mappings from $\mathbb R^n$ to $\mathbb R$ which quantify how much energy of a vector is packed into only a few components. On a normalized scale, the sparsest possible vector (only a single component is non-zero) should have a sparseness of one, whereas a vector with all elements equal should have a sparseness of zero. 

In this paper, we use a sparseness measure based on the relationship between the $L_1$ norm and the $L_2$ norm:
\beq
\mbox{sparseness}(\x) = \frac{\sqrt{n} - \left(\sum |x_i|\right)/\sqrt{\sum x_i^2}}{  \sqrt{n} - 1 },
\eeq
where $n$ is the dimensionality of $\x$. This function evaluates to unity if and only if $\x$ contains only a single non-zero component, and takes a value of zero if and only if all components are equal (up to signs), interpolating smoothly between the two extremes.

\subsection{NMF with sparseness constraints}

Our aim is to constrain NMF to find solutions with desired degrees of sparseness. The first question to answer is then: what exactly should be sparse? The basis vectors $\W$ or the coefficients $\bH$? This is a question that cannot be given a general answer; it all depends on the specific application in question. Further, just transposing the data matrix switches the role of the two, so it is easy to see that the choice of which to constrain (or both, or none) must be made by the experimenter. 

For example, a doctor analyzing disease patterns might assume that most diseases are rare (hence sparse) but that each disease can cause a large number of symptoms. Assuming that symptoms make up the rows of her matrix and the columns denote different individuals, in this case it is the `coefficients' which should be sparse and the `basis vectors' unconstrained. On the other hand, when trying to learn useful features from a database of images, it might make sense to require both $\W$ and $\bH$ to be sparse, signifying that any given object is \emph{present} in few images and \emph{affects} only a small part of the image.

These considerations lead us to defining NMF with sparseness constraints as follows:\\ 
\rule{\textwidth}{0.6mm}\\[0mm]
{\sffamily
{\bfseries Definition:} NMF with sparseness constraints\\

Given a non-negative data matrix $\V$ of size $N \times T$, find the non-negative matrices $\W$ and $\bH$ of sizes $N \times M$ and $M \times T$ (respectively) such that 
\beq
E(\W,\bH) = \|\V - \W\bH\|^2
\label{eq:obj}
\eeq
is minimized, under \emph{optional constraints} 
\begin{eqnarray}
\mbox{sparseness}(\wi) & = & S_w, \;\; \forall i \\
\mbox{sparseness}(\hi) & = & S_h, \;\; \forall i,
\end{eqnarray}
where $\wi$ is the $i$:th \emph{column} of $\W$ and $\hi$ is the $i$:th \emph{row} of $\bH$. Here, $M$ denotes the number of components, and $S_w$ and $S_h$ are the desired sparsenesses of $\W$ and $\bH$ (respectively). These three parameters are set by the user.}\\[-1mm]
\rule{\textwidth}{0.6mm} 

Note that we did not constrain the scales of $\wi$ or $\hi$ yet. However, since $\wi\hi = (\wi\lambda)(\hi/\lambda)$ we are free to arbitrarily fix any norm of either one. In our algorithm, we thus choose to fix the $L_2$ norm of $\hi$ to unity, as a matter of convenience.

\subsection{Algorithm}
\label{sec:algo}

We have devised a projected gradient descent algorithm for NMF with sparseness constraints. This algorithm essentially takes a step in the direction of the negative gradient, and subsequently projects onto the constraint space, making sure that the taken step is small enough that the objective function (\ref{eq:obj}) is reduced at every step. The main muscle of the algorithm is the projection operator which enforces the desired degree of sparseness. This operator is described in detail following this algorithm. \\[2mm]
\rule{\textwidth}{0.6mm}\\[0mm]
{\sffamily
{\bfseries Algorithm:} NMF with sparseness constraints

\begin{enumerate}
\item Initialize $\W$ and $\bH$ to random positive matrices
\item If sparseness constraints on $\W$ apply, then project each column of $\W$ to be non-negative, have unchanged $L_2$ norm, but $L_1$ norm set to achieve desired sparseness
\item If sparseness constraints on $\bH$ apply, then project each row of $\bH$ to be non-negative, have unit $L_2$ norm, and $L_1$ norm set to achieve desired sparseness
\item Iterate
\begin{enumerate}
\item If sparseness constraints on $\W$ apply, 
\begin{enumerate}
\item Set $\W := \W - \mu_{\W}(\W\bH - \V)\bH^T$
\item Project each column of $\W$ to be non-negative, have unchanged $L_2$ norm, but $L_1$ norm set to achieve desired sparseness
\end{enumerate}
else take standard multiplicative step $\W := \W \otimes (\V\bH^T) \oslash (\W\bH\bH^T)$
\item If sparseness constraints on $\bH$ apply, 
\begin{enumerate}
\item Set $\bH := \bH - \mu_{\bH}\W^T(\W\bH - \V)$
\item Project each row of $\bH$ to be non-negative, have unit $L_2$ norm, and $L_1$ norm set to achieve desired sparseness
\end{enumerate}
else take standard multiplicative step $\bH := \bH \otimes (\W^T\V) \oslash (\W^T\W\bH)$
\end{enumerate}
\end{enumerate}
Above, $\otimes$ and $\oslash$ denote elementwise multiplication and division, respectively. Moreover, $\mu_{\W}$ and $\mu_{\bH}$ are small positive constants (stepsizes) which must be set appropriately for the algorithm to work. Fortunately, they need not be set by the user; our implementation of the algorithm automatically adapts these parameters. The multiplicative steps are directly taken from \cite{LeeDD01} and are used when constraints are not to be applied.\\[-1mm]}
\rule{\textwidth}{0.6mm}\\ 

Many of the steps in the above algorithm require a projection operator which enforces sparseness by explicitly setting both $L_1$ and $L_2$ norms (and enforcing non-negativity). This operator is defined as follows\\[2mm]
\rule{\textwidth}{0.6mm}\\[-7mm]
{\sffamily 
\begin{description}

\item[problem] Given any vector $\x$, find the closest (in the euclidean sense) \emph{non-negative} vector $\s$ with a given $L_1$ norm and a given $L_2$ norm.

\item[algorithm] The following algorithm solves the above problem. See below for comments.

\begin{enumerate}
\item Set $s_i := x_i + (L_1-\sum x_i)/\mbox{dim}(\x), \;\;\forall i$
\item Set $Z :=  \{ \}$
\item Iterate
\begin{enumerate}
\item Set $m_i := \left\{ \begin{array}{ll}
L_1/(\mbox{dim}(\x) - \mbox{size}(Z)) & \mbox{if\;\;} i\notin Z \\
0 & \mbox{if\;\;} i\in Z 
\end{array}
\right.
$

\item Set $\s := \m + \alpha(\s-\m)$, where $\alpha \geq 0$ is selected such that the resulting $\s$ satisfies the $L_2$ norm constraint. This requires solving a quadratic equation.

\item If all components of $\s$ are non-negative, return $\s$, end

\item Set $Z := Z \cup \{ i; s_i<0 \}$
\item Set $s_i := 0,\;\; \forall i\in Z$
\item Calculate $c := (\sum s_i - L_1)/(\mbox{dim}(\x) - \mbox{size}(Z))$
\item Set $s_i := s_i - c, \;\; \forall i\notin Z$
\item Go to (a)
\end{enumerate} 
\end{enumerate}

\end{description}
}
\vspace{-3mm}
\rule{\textwidth}{0.6mm}\\ 

In words, the above algorithm works as follows: We start by projecting the given vector onto the hyperplane $\sum s_i = L_1$. Next, within this space, we project to the closest point on the joint constraint hypersphere (intersection of the sum and the $L_2$ constraints). This is done by moving radially outward from the center of the sphere (the center is given by the point where all components have equal values). If the result is completely non-negative, we have arrived at our destination. If not, those components that attained negative values must be fixed at zero, and a new point found in a similar fashion under those additional constraints. 

Note that, once we have a solution to the above \emph{non-negative} problem, it would be straightforward to extend it to a general solution without non-negativity constraints. If a given component of $\x$ is positive (negative), we know because of the symmetries of $L_1$ and $L_2$ norms that the optimal solution $\s$ will have the corresponding component positive or zero (negative or zero). Thus, we may simply record the signs of $\x$, take the absolute value, perform the projection in the first quadrant using the algorithm above, and re-enter the signs into the solution.

In principle, the devised projection algorithm may take as many as $\mbox{dim}(\x)$ iterations to converge to the correct solution (because at each iteration the algorithm either converges, or at least one component is added to the set of zero valued components). In practice, however, the algorithm converges much faster. In section~\ref{sec:experiments} we show that even for extremely high dimensions the algorithm typically converges in only a few iterations.

\subsection{Matlab implementation}

Our software package, available at \texttt{http://www.cs.helsinki.fi/patrik.hoyer/} implements all the details of the above algorithm. In particular, we monitor the objective function $E$ throughout the optimization, and adapt the stepsizes to ensure convergence. The software package contains, in addition to the projection operator and NMF code, all the files needed to reproduce the results described in this paper, with the exception of datasets. For copyright reasons the face image databases are not included, but they can easily be downloaded separately from their respective www addresses.


\section{Experiments with sparseness constraints}
\label{sec:experiments}

In this section, we show that adding sparseness constraints to NMF can make it find parts-based representations in cases where unconstrained NMF does not. In addition, we experimentally verify our claim that the projection operator described in Section~\ref{sec:algo} converges in only a few iterations even when the dimensionality of the vector is high.

\subsection{Representations learned from face image databases}

Recall from Section~\ref{sec:nmf} the mixed results of applying standard NMF to face image data. Lee and Seung \cite{LeeDD99} originally showed that NMF found a parts-based representation when trained on data from the CBCL database. However, when applied to the ORL dataset, in which images are not as well aligned, a global decomposition emerges. These results were shown in Figure~\ref{fig:nmf}a and \ref{fig:nmf}b. To compare, we applied sparseness constrained NMF to both face image datasets. 

\begin{figure}
\centerline{
\resizebox{120mm}{!}{
\includegraphics{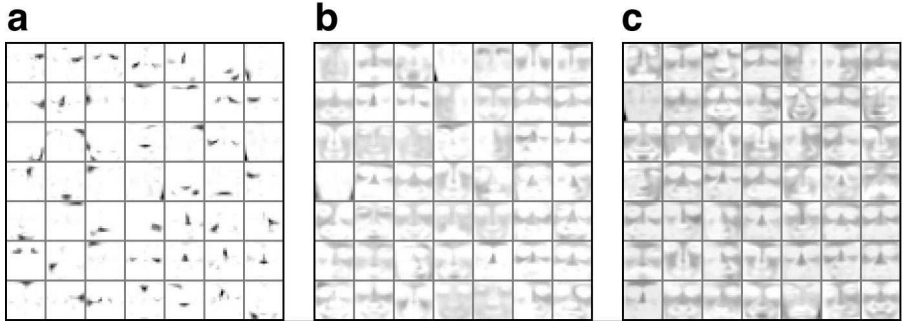}}}
\caption{Features learned from the CBCL face image database using NMF with sparseness constraints. {\bf (a)} The sparseness of the basis images were fixed to 0.8, slightly higher than the average sparseness produced by standard NMF, yielding a similar result. The sparseness of the coefficients was unconstrained. {\bf (b)} Here, we switched the sparseness constraints such that the coefficients were constrained to 0.8 but the basis images were unconstrained. Note that this creates a global representation similar to that given by vector quantization \cite{LeeDD99}. {\bf (c)} Illustration of another way to obtain a global representation: setting the sparseness of the basis images to a low value (here: 0.2) also yields a non-local representation.
\label{fig:nmfsccbcl}}
\end{figure}

For the CBCL data, some resulting bases are shown in Figure~\ref{fig:nmfsccbcl}. Setting a high sparseness value for the basis images results in a local representation similar to that found by standard NMF. However, we want to emphasize the fact that sparseness constrained NMF does not always lead to local solutions: Global solutions can be obtained by deliberately setting a low sparseness on the basis images, or by requiring a high sparseness on the coefficients (forcing each coefficient to try to represent more of the image).

\begin{figure}
\centerline{
\resizebox{120mm}{!}{
\includegraphics{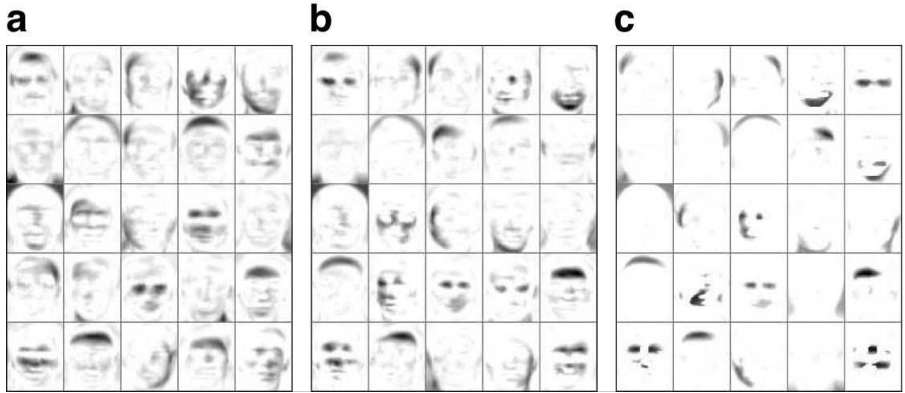}}}
\caption{Features learned from the ORL face image database using NMF with sparseness constraints. When increasing the sparseness of the basis images, the representation switches from a global one (like the one given by standard NMF, cf Figure~\ref{fig:nmf}b) to a local one. Sparseness levels were set to {\bf (a)} 0.5 {\bf (b)} 0.6 {\bf (c)} 0.75. 
\label{fig:nmfscorl}}
\end{figure}

The ORL database provides the more interesting test of the method. In Figure~\ref{fig:nmfscorl} we show bases learned by sparseness constrained NMF, for various sparseness settings. Note that our method can learn a parts-based representation of this dataset, in contrast to standard NMF. Also note that the representation is not very sensitive to the specific sparseness level chosen. 

\subsection{Basis derived from natural image patches}

\begin{figure}
\centerline{
\resizebox{140mm}{!}{
\includegraphics{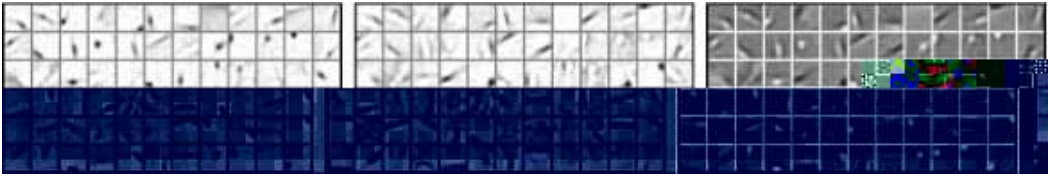}}}
\caption{Basis vectors from ON/OFF-filtered natural images obtained using NMF with sparseness constraints. The sparseness of the coefficients was fixed at 0.85, and the sparseness of the basis images was unconstrained. As opposed to standard NMF (cf Figure~\ref{fig:nmf}c), the representation is based on oriented, Gabor-like, features.
\label{fig:nmfscnat}}
\end{figure}

In Figure~\ref{fig:nmf}c we showed that standard NMF applied to natural image data produces only circular features, not oriented features like those employed by modern image processing techniques. Here, we tested the result of using additional sparseness constraints. Figure~\ref{fig:nmfscnat} shows the basis vectors obtained by putting a sparseness constraint on the coefficients ($S_h = 0.85$) but leaving the sparseness of the basis vectors unconstrained. In this case, NMF learns oriented features that represent edges and lines. Such oriented features are widely regarded as the best type of low-level features for representing natural images, and similar features are also used by the early visual system of the biological brain \cite{Field87,Simoncelli92,Olshausen96b,Bell97a}. This example illustrates that sparseness constrained NMF does not simply `sparsify' the result of standard, unconstrained NMF, but rather can find qualitatively different parts-based representations that are more compatible with the sparseness assumptions.

\subsection{Convergence of algorithm implementing the projection step}

\begin{figure}
\centerline{
\resizebox{60mm}{!}{
\includegraphics{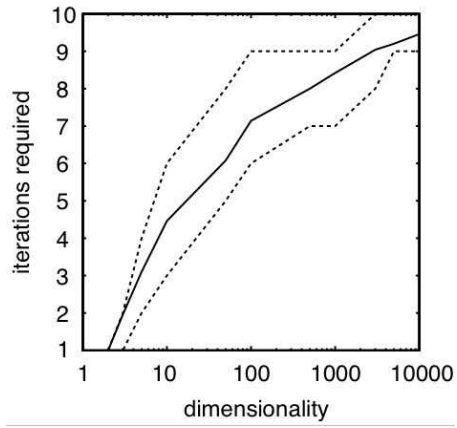}}}
\caption{Number of iterations required for the projection algorithm to converge, in the worst-case scenario tested (desired sparseness 0.9, initial sparseness 0.1). The solid line shows the average number (over identical random trials) of iterations required, the dashed lines show the minimum and maximum iterations. Note that the number of iterations grows very slowly with the dimensionality of the problem.
\label{fig:projtests}}
\end{figure}

To verify the performance of our projection method we performed extensive tests, varying the number of dimensions, the desired degree of sparseness, and the sparseness of the original vector. The desired and the initial degrees of sparseness were set to 0.1, 0.3, 0.5, 0.7, and 0.9, and the dimensionality of the problem was set to 2, 3, 5, 10, 50, 100, 500, 1000, 3000, 5000, and 10000. All combinations of sparsenesses and dimensionalities were analyzed. Based on this analysis, the worst case (most iterations on average required) was when the desired degree of sparseness was high (0.9) but the initial sparseness was low (0.1). In Figure~\ref{fig:projtests} we plots the number of iterations required for this worst case, as a function of dimensionality. Even in this worst-case scenario, and even for the highest tested dimensionality, the algorithm never required more than 10 iterations to converge. Thus, although we do not have analytical bounds on the performance on the algorithm, empirically the projection method performs extremely well.

\section{Relation to other recent work}
\label{sec:discussion}

\subsection{Extensions of NMF}

Several authors have noted the shortcomings of standard NMF, and suggested extensions and modifications of the original model. Li et al \cite{LiS01} noted that NMF found only global features from the ORL database (see Figure~\ref{fig:nmf}b) and suggested an extension they call \emph{Local} Non-negative Matrix Factorization (LNMF).  Their method indeed produces local features from the ORL database, similar to those given by our method (Figure~\ref{fig:nmfscorl}c). However, it does not produce oriented filters from natural image data (results not shown). Further, there is no way to explicitly control the sparseness of the representation, should this be needed.

In \cite{Hoyer02NNSP} the current author extended the NMF framework to include an adjustable sparseness parameter. The present paper is an extension of those ideas. The main improvement is that in the present model sparseness is adjusted explicitly, rather than implicitly. This means that one does not any more need to employ trial-and-error to find the parameter setting that yields the desired level of sparseness. 

Finally, Liu et al \cite{Liu03} also noted the need for incorporating the notion of sparseness, and suggested an extension termed \emph{Sparse} Non-negative Matrix Factorization (SNMF). Their extension is similar in spirit and form to that given in \cite{Hoyer02NNSP} with the added benefit of yielding a more convenient, faster algorithm. Nevertheless, it also suffers from the drawback that sparseness is only controlled implicitly. Furthermore, their method does not yield oriented features from natural image data (results not shown).

In summary, the framework presented in the present paper improves on these previous extensions by allowing explicit control of the statistical properties of the representation. 

In order to facilitate the use of, and comparison between, the various extensions of NMF, they are all provided as part of the Matlab code package distributed with this paper. Using this package readers can effortlessly verify our current claims by applying the algorithms to the various datasets. Moreover, the methods can be compared head-to-head on new interesting datasets.

\subsection{Non-negative independent component analysis}

Our method has a close connection to the statistical technique called independent component analysis (ICA) \cite{Hyva01book}. ICA attempts to find a matrix factorization similar to ours, but with two important differences. First, the signs of the components are in general not restricted; in fact, symmetry is often assumed, implying an approximately equal number of positive and negative elements. Second, the sources are not forced to any desired degree of sparseness (as in our method) but rather sparseness is incorporated into the objective function to be optimized. The sparseness goal can be put on either $\W$ or $\bH$, or both \cite{Stone02}.

Recently, some authors have considered estimating the ICA model in the case of one-sided,  non-negative sources \cite{Plumbley03,Oja04}. In these methods, non-negativity is not specified as a constraint but rather as an objective; hence, complete non-negativity of the representation is seldom achieved for real-life datasets. Nevertheless, one can show that if the linear ICA model holds, with non-negative components, these methods can identify the model.


\section{Conclusions}
\label{sec:conclusions}

Non-negative matrix factorization (NMF) has proven itself a useful tool in the analysis of a diverse range of data. One of its most useful properties is that the resulting decompositions are often intuitive and easy to interpret because they are sparse. Sometimes, however, the sparseness achieved by NMF is not enough; in such situations it might be useful to control the degree of sparseness explicitly. Our main contributions of this paper were (a) to describe a projection operator capable of simultaneously enforcing both $L_1$ and $L_2$ norms and hence any desired degree of sparseness, (b) to show its use in the NMF framework for learning representations that could not be obtained by regular NMF, and (c) to provide a software package to enable researchers and practitioners to easily perform NMF and its various extensions. We hope that all three contributions will prove useful to the field of data-analysis.


\bibliographystyle{unsrt} 
\bibliography{/users/phoyer/bib/collection,/users/phoyer/bib/personal,/users/phoyer/bib/others}

\section*{Acknowledgements}
The author wishes to thank Jarmo Hurri, Aapo Hyv\"arinen, and Fabian Theis for useful discussions and comments on the manuscript.

\end{document}